%% file: main.tex
\documentclass[10pt,twocolumn,letterpaper]{article}

\usepackage[pagenumbers]{cvpr} 

\input{preamble}

\input{config}

\definecolor{cvprblue}{rgb}{0.21,0.49,0.74}
\usepackage[pagebackref,breaklinks,colorlinks,citecolor=cvprblue]{hyperref}

\def\paperID{14831} 
\def\confName{CVPR}
\def\confYear{2024}

\title{ Feedback RoI Features Improve Aerial Object Detection}

\author{
Botao Ren$^1$
\and
Botian Xu$^1$
\and
Tengyu Liu$^2$
\and
Jingyi Wang$^1$
\and
Zhidong Deng$^1$\thanks{Corresponding author. Zhidong Deng is with Beijing National Research Center for Information Science and Technology (BNRist), Institute for Artificial Intelligence at Tsinghua University (THUAI), Department of Computer Science, State Key Laboratory of Intelligent Technology and Systems, Tsinghua University, Beijing 100084, China.}
\and
\ \\ $^1$Tsinghua University\\ $^2$National Key Laboratory of General Artificial Intelligence, BIGAI
}

\begin{document}
\maketitle
\input{sec/0_abstract}    
\input{sec/1_intro}
\input{sec/2_relatedworks}

\input{sec/3_method}
\input{sec/4_experiments}
\input{sec/5_conclusion}
{
    \small
    \bibliographystyle{ieeenat_fullname}
    \bibliography{main}
}


\end{document}

%% file: preamble.tex
\usepackage[dvipsnames]{xcolor}


%% file: config.tex
\usepackage{cvpr}
\usepackage{amsmath}
\usepackage{graphicx}
\usepackage{adjustbox}
\usepackage{amssymb}
\usepackage{subcaption}
\usepackage{svg}
\usepackage[utf8]{inputenc} 
\usepackage[T1]{fontenc}    
\usepackage{url}            
\usepackage{booktabs}       
\usepackage{amsfonts}       
\usepackage{nicefrac}       
\usepackage{microtype}      
\usepackage{xcolor}         
\usepackage{xspace}
\usepackage{soul}

\usepackage{acronym}

\usepackage{ifthen}
\usepackage{color}
\usepackage{tabularx}
\usepackage{multirow}

\newboolean{flag}
\setboolean{flag}{true} 
\setboolean{flag}{false} 
\usepackage{stfloats}

\newcommand{\specialtext}[1]{%
  \ifthenelse{\boolean{flag}}{%
    \textcolor{purple}{#1}%
  }{%
    #1%
  }%
}

\definecolor{ORANGE}{RGB}{194,115,44}
\definecolor{BLUE}{RGB}{124,173,184}
\definecolor{red}{RGB}{255,0,0}
\definecolor{blue}{RGB}{0,0,255}

\acrodef{fpn}[FPN]{Feature Pyramid Network}
\acrodef{roi}[RoI]{Region of Interest}

\newcommand{\model}[0]{\texttt{Flex}\xspace}

\newcommand{\highlight}[1]{\textcolor{blue}{\textbf{#1}}}

%% file: sec/0_abstract.tex
\begin{abstract}
Neuroscience studies have shown that the human visual system utilizes high-level feedback information to guide lower-level perception, enabling adaptation to signals of different characteristics. In light of this, we propose \underline{F}eedback multi-\underline{L}evel feature \underline{Ex}tractor (\model) to incorporate a similar mechanism for object detection. \model refines feature selection based on image-wise and instance-level feedback information in response to image quality variation and classification uncertainty. Experimental results show that \model offers consistent improvement to a range of existing SOTA methods on the challenging aerial object detection datasets including DOTA-v1.0, DOTA-v1.5, and HRSC2016. \specialtext{Although the design originates in aerial image detection, further experiments on MS COCO also reveal our module's efficacy in general detection models.} Quantitative and qualitative analyses indicate that the improvements are closely related to image qualities, which match our motivation. 
\end{abstract}

%% file: sec/1_intro.tex
\section{Introduction}
\label{sec:intro}

Despite impressive advances in object detection \cite{redmon2016you, redmon2018yolov3, liu2016ssd, ren2015faster}, detecting objects in aerial images remains a challenging problem for two reasons: (a) the objects vary drastically in orientation, scale, aspect ratios, and density; (b) the qualities of images vary a lot due to difference in various factor including sensor condition, lighting, and air quality. Most existing methods address the former challenge with advanced neural network architectures \cite{han2021redet, xie2021oriented, yang2019scrdet} and objectives \cite{yang2021learning, yang2021rethinking, yang2022kfiou}, and delegate the latter challenge with data augmentation. 

However, research in neuroscience~\cite{webster2002neural, gilbert2007brain, friston2010free} has established that the human visual system has a special mechanism called \textit{neural adaptation} to handle low-quality visual signals efficiently. Inspired by this previously unexplored mechanism, we propose a plug-and-play feedback-refine module that adaptively fuses multi-level features.

Our module is compatible with all two-stage detectors, in which the first stage generates a multi-level feature map with an \ac{fpn} \cite{lin2017feature} and a set of \ac{roi} \cite{girshick2014rich} proposals, and the second stage extracts features from them for bounding box and class prediction. Specifically, after the first pass of detection, for each detection, our module predicts a set of feature fusion weights based on the image-level features, the classification results, and the area of the \ac{roi}. The first two features reflect the ambiguities in the image, and the last provides an anchor for the fusion weights. We then fuse the \ac{fpn} features from the first stage with the new weights for the final classification.

Through extensive experiments, we show that including our plug-and-play module significantly improves recent two-stage object detection models on challenging aerial object detection datasets DOTA-v1.0, DOTA-v1.5\cite{xia2018dota}, and HRSC2016\cite{liu2017high}. \specialtext{To verify that our module is generally applicable and can bring consistent improvements, we also conducted experiments on MS COCO\cite{lin2014microsoft}.} We further perform detailed analyses to unveil that the improvements indeed come from adapting to images of different qualities. The contribution can be summarized as three-fold:
\specialtext{
\begin{itemize}
    \item We propose a simple yet effective module to leverage feedback information to improve detection in aerial images. It adaptively fuses multi-level features to address the challenges of varying object characteristics and image quality in aerial images. 
    \item With \model we verify and validate how the challenges affect detection performance, and how image- and object-level information could be useful.
    \item Experimental results demonstrate that our feedback-based module can be plugged into existing SOTA methods and provides consistent improvements on challenging aerial object detection datasets including DOTA-v1.0, DOTA-v1.5, and HRSC2016. It also offers notable performance gains on MS COCO when images become blurred.
\end{itemize}
}

%% file: sec/2_relatedworks.tex
\section{Related Works}
\label{sec:relatedworks}

\subsection{Object Detection in Aerial Images}

Object detection in aerial imagery presents unique challenges due to the scale, orientation, and aspect ratio variations of objects, as well as dense clusters of small objects and variances in image quality, including lighting and air quality. Recent advances, regardless of single-stage and two-stage methods, have focused on addressing the former challenge. Representative two-stage methods for aerial images include ReDet\cite{han2021redet}, Oriented RCNN\cite{xie2021oriented}, SCRDet\cite{yang2019scrdet}, SASM\cite{hou2022shape}, Gliding vertex\cite{xu2020gliding}, \ac{roi} Transformer\cite{ding2019learning}, etc., while single-stage methods include R\textsuperscript{3}Det\cite{yang2021r3det}, S\textsuperscript{2}ANet\cite{han2021align}, DAL\cite{ming2021dynamic}, etc. ReDet\cite{han2021redet} and ARC\cite{pu2023adaptive} modified the convolution layers to explicitly cope with rotation; Oriented RCNN\cite{xie2021oriented} designed a better representation of oriented bounding boxes; to deal with objects with large aspect ratios, Hou et al. designed A\ac{fpn}\cite{hou2022refined} where kernels can have asymmetric shapes. S\textsuperscript{2}ANet\cite{han2021align} aligns features between axis-aligned convolutional features and oriented objects. Novel loss functions such as GWD\cite{yang2021rethinking}, KLD\cite{yang2021learning}, and KFIoU\cite{yang2022kfiou} were proposed to locate and regress oriented bounding boxes more accurately. In addition, multi-scale training and testing strategies are applied to detect small and dense objects. Our method stands as the first attempt in addressing the challenge of varying image qualities. 

\subsection{Multi-level Feature Fusion}

Multi-level feature fusion has been widely used in object detection to detect objects of various scales. Bell et al. \cite{bell2016inside} proposed ION, which connects \ac{roi} features of different granularity to form new features, and then classifies and regresses \ac{roi}. RON\cite{kong2017ron} uses a reverse connection to pass the coarse-grained information of the latter layer to the lower layer. Lin et al.\cite{lin2017feature} introduced Feature Pyramid Network (\ac{fpn}), which builds a hierarchy of feature maps via top-down, bottom-up, and lateral connections, enabling the network to better leverage information from different scales. To further improve the effectiveness of \ac{fpn} for object detection, several recent works have proposed approaches that adaptively mix multi-level features. Gong et al.\cite{gong2021effective} proposed to adjust the fusion factor in \ac{fpn} to make it more suitable for detecting tiny objects. ASFF\cite{liu2019learning} fuses feature maps from different layers with learnable weights, showing advantages over element-wise sum or concatenation. SENet\cite{hu2018squeeze} and CBAM\cite{woo2018cbam} adaptively modulate channel-wise information. Zhao et al. proposed M2Det\cite{zhao2019m2det}, which generates and extracts multi-scale features and then synthesizes a multi-level feature pyramid using channel attention. Closest to our work, Zhen et al.\cite{zhen2023towards} propose to use adaptive multi-layer feature fusion in aerial images at the channel level.

Our work is motivated by the observation that how we fuse multi-level features should depend on the image quality and the proposed \ac{roi}. 
Although adaptive in some sense, prior works may fail to capture the interplay between image-level information (e.g., lighting and quality) and object-level semantics, which we address by providing such information as feedback to readjust the fusion process.

\subsection{Adaptive Feedback in Neural Networks}

In recent years, there has been growing interest in incorporating feedback mechanisms in neural networks for various tasks, including object detection. These mechanisms are inspired by the feedback processes in the human visual system, where higher-level visual areas send signals back to lower-level areas to refine feature representations. Cao et al.\cite{cao2015look} proposed a gating strategy to change the activation state of hidden layers based on the optimization goals of the network. Zamir et al.\cite{zamir2017feedback} proposed Feedback Networks where the representation is formed iteratively with feedback received from the previous iteration's output. Li et al.\cite{li2019feedback} used high-level information to refine low-level representations for image super-resolution. Feng et al.\cite{feng2019attentive} designed an attentive feedback module for Boundary-Aware Salient Object Detection (BSD), where the predictions from the previous stage are fed to the latter stage to implement corrections. Huang et al.\cite{huang2020neural} proposed CNN-F (CNN with Feedback), which introduces generative feedback with latent variables to improve performance.

Inspired by this mechanism, our method uses image-level and \ac{roi} features to condition the feature fusion process in the second stage, effectively refining the representation obtained.

%% file: sec/3_method.tex
\section{Methodology}


\begin{figure*}[htbp] 
  \centering
  \includegraphics[width=\textwidth]{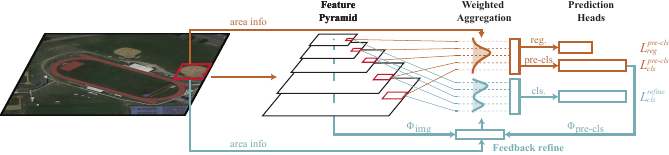}
  \caption{Overview of the \model. In the \textcolor{ORANGE}{\textbf{pre-classification}} stage, \model obtains \ac{roi} features $f_{\text{pre}}$ by fusing different pyramid levels with weights given by a Gaussian kernel and then computes the bounding box and class prediction. The \textcolor{BLUE}{\textbf{feedback refine}} stage then computes (1) image content feedback $\Phi^{\text{img}}$ from the feature pyramid and (2) classification feedback $\Phi^{\text{cls}}$ for each \ac{roi} from the pre-classification result. Then we use $\Phi^{\text{img}}$, $\Phi^{\text{cls}}$ and \ac{roi} area to compute refined feature weights with which the final feature $f_{\text{refine}}$ is obtained and used to give the final result with a second detection module. The whole model is jointly trained with standard loss functions for object detection.}
  \label{arch_overview}
\end{figure*}

Motivated by the neural adaptation mechanism, we designed \model, \underline{F}eedback Multi-\underline{L}evel \ac{roi} Feature \underline{Ex}tractor, a feedback refine module that plugs into any two-stage detection models. We name the original two-stage detection as the \textcolor{ORANGE}{\textbf{pre-classification}} stage and our design as the \textcolor{BLUE}{\textbf{feedback refine}} stage. \model takes the image feature from the \ac{fpn}, the classification logits from the \textcolor{ORANGE}{\textbf{pre-classification}} stage, and the area of an \ac{roi} to predict a set of fusion weights. We then use these weights to aggregate \ac{fpn} features and classify the \ac{roi} in the \textcolor{BLUE}{\textbf{feedback refine}} stage. We outline \model in \cref{arch_overview}.

\subsection{The Pre-Classification Stage}
\model accepts any two-stage object detection method with a feature pyramid as its pre-classification stage, with one slight modification on feature extraction. 

The original way of extracting the feature for an \ac{roi} from an $N$-level feature pyramid $f=\{f_k\}_{k=1}^N$ is to directly take the feature from the $i$-th level, where

\begin{equation}
    i = \max(\min(\lfloor 1 + \log_2{\sqrt{wh}/\delta} \rfloor, N), 1)
\end{equation}
is determined by the \ac{roi} area, $w$ and $h$ are the \ac{roi}'s height and width. $\delta$ is a scale factor which we set to $56$ following the original publication.

To better leverage information from multiple levels, we do not restrict feature extraction to be from a single level. Instead, we use a Gaussian kernel centered at $i$ to obtain a set of weights for the pre-classification stage by first dropping the floor operation:
\begin{equation} \label{eq:target_level}
    i = \max(\min(1 + \log_2{\sqrt{wh}/\delta}, N), 1),
\end{equation}
and then compute
\begin{equation} \label{eq:gaussian_weight}
    W_k=\frac{1}{\sigma \sqrt{2 \pi}} e^{-\frac{(k-i)^2}{2 \sigma^2}}, k \in[1, N]
\end{equation}
where $W_k$ is the weight of the $k$-th level, and $\sigma$ is a hyperparameter.

We then compute a weighted sum of $f$ as the \ac{roi} feature in the pre-classification stage:

\begin{equation} \label{sumupfeature}
    f_{\text{pre}}=\frac{\sum_{k=1}^N W_k f_k}{\sum_{j=1}^N W_j}. 
\end{equation}


\subsection{Computing Feedbacks from Pre-Classification}
We then compute the image and classification feedback from the results of the pre-classification stage.

We obtain the image feedback $\Phi^{\text{img}}\in\mathbb{R}^N$ from the $N$-level \ac{fpn} features $f$, whose $i$-th level feature $f_i$ has size $[C_0, H_0/2^i, W_0/2^i]$, $C_0$, $H_0$ and $W_0$ being the channel size, height, and width of the lowest level feature map. We project each feature level to the same height and width $[C_i,H_0/2^N ,W_0/2^N ]$ by a convolutional network, where
\begin{equation}
    C_i = C_0 / 2^{N+1-\min{(2, i)}}
\end{equation}
Then, we concatenate the features channel-wise and feed them through another convolutional network to obtain $\Phi^{\text{img}}$. Notice the shape of $\Phi^\text{img}$ since we need one value for each feature level. 

We further obtain the classification feedback $\Phi^\text{cls}\in\mathbb{R}^{2\lfloor N/2\rfloor +1}$ for each \ac{roi} by applying a fully connected network to the classification output of the \ac{roi} in the pre-classification stage. We require an odd number of values to perform the interpolation in the feedback refine stage. 

\subsection{Feedback Refine: Multi-Level Feature Extraction}


\begin{figure*}[htbp]
    \centering
    \includegraphics[width=0.6\textwidth]{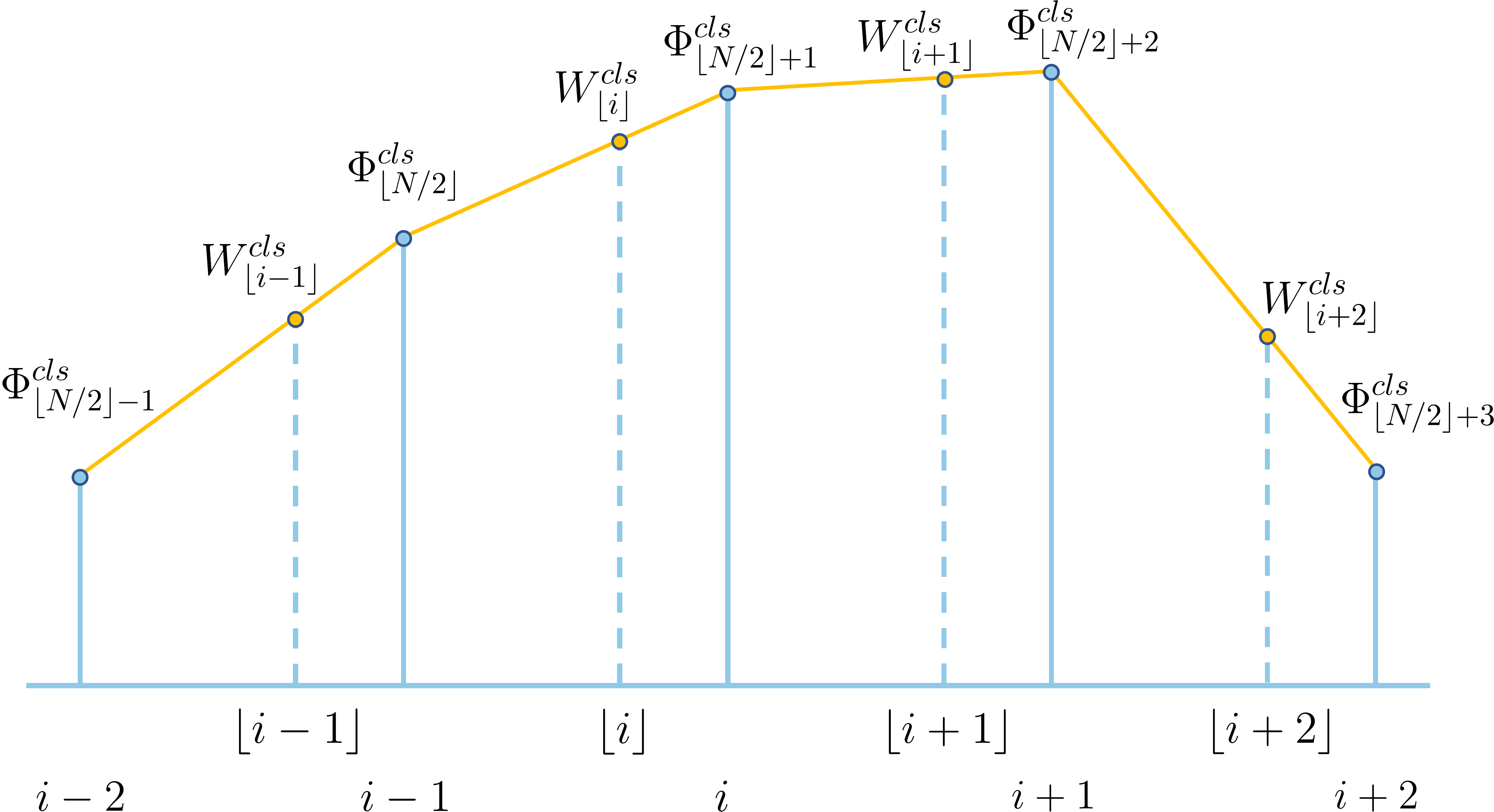}
    \caption{\ac{roi} area-aware weight calculation from classification feedback $\Phi^{\text{cls}}$. We interpret $\Phi^{\text{cls}}\in R^{2\lfloor N/2\rfloor +1}$ as a piece-wise linear kernel centered at the target level $i$ which is determined by \ac{roi} area according to \cref{eq:target_level}. $W_k^{\text{cls}}$ is then evaluated at $k$ with linear interpolation.}
    \label{interpolation}
\end{figure*}



In the feedback refinement stage, we compute the feature weight as 
\begin{equation}
W_k^{FB} = \Phi_k^{\text{img}} W_k^{\text{cls}}, 
\end{equation}
where $\Phi_k^{\text{img}}$ is the $k$-th value of $\Phi^\text{img}$. To obtain \specialtext{$W_k^{\text{cls}}\in \mathbb{R}^N$}, we interpret $\Phi_k^{\text{cls}}$ as a piece-wise linear kernel centered at the target level $i$ such that
\begin{equation}
    W_k^{\text{cls}}=\Phi_{\lfloor j \rfloor}^{\text{cls}} + (j - \lfloor j \rfloor)(\Phi_{\lfloor j \rfloor +1}^{\text{cls}} - \Phi_{\lfloor j \rfloor}^{\text{cls}})
\end{equation}
for $i-\left\lfloor \frac{N}{2} \right\rfloor \leq k \leq i+\left\lfloor \frac{N}{2} \right\rfloor$ and $0$ otherwise. Here $j = k-i + \lfloor \frac{N}{2} \rfloor$, and $\Phi^\text{cls}_k$ is the $k$-th value of $\Phi^\text{cls}$.
We discard feature levels that are too far from $i$, i.e., not in the range of $[i-\lfloor N/2\rfloor, i+ \lfloor N/2 \rfloor ]$. We visually illustrate the calculation in \cref{interpolation}. We remark that directly learning a weight vector $\Phi^{\text{cls}}$ of dimension $N$ out of the class prediction is also viable but fails to leverage the area information.

We again normalize $W^{FB}$ to obtain the feedback-refined \ac{roi} feature:

\begin{equation}
    f_{\text {refine}}=\frac{\sum_{i=1}^N W_i^{FB} f_i}{\sum_{j=1}^N W_j^{FB}}.
\end{equation}
and $f_\text{refine}$ is used for the final classification for each \ac{roi}.

\subsection{Losses and Hyperparameters}

The feedback refine stage is trained with the pre-classification stage in an end-to-end manner following the overall loss function:
\begin{equation}
    L =  L_\text{reg}^\text{pre-cls} + \gamma L_{\text{cls}}^\text{pre-cls} + L_{\text{cls}}^\text{refine}.
\end{equation}

We use the standard bounding box regression loss $L_\text{reg}$ and cross-entropy classification loss $L_{\text{cls}}$ for both stages. $\gamma$ is a hyperparameter.

%% file: sec/4_experiments.tex
\section{Experiment}
\subsection{Experimental Setup}
\subsubsection{Dataset}

DOTA\cite{xia2018dota} is a large-scale dataset for object detection in aerial images. 

DOTA-v1.0 contains 2,806 images and 188,282 instances and is labeled with 15 categories, annotated as: \texttt{Plane} (PL), \texttt{Baseball Diamond} (BD), \texttt{Bridge} (BR), \texttt{Ground Track Field} (GTF), \texttt{Small Vehicle} (SV), \texttt{Large Vehicle} (LV), \texttt{Ship} (SH), \texttt{Tennis Court} (TC), \texttt{Basketball Court} (BC), \texttt{Storage Tank} (ST), \texttt{Soccer Ball Field} (SBF), \texttt{Roundabout} (RA), \texttt{Harbor} (HA), \texttt{Swimming Pool} (SP), and \texttt{Helicopter} (HC). The images range in size from 800×800 to 4000×4000 and contain targets with various scales, orientations, and shapes. Following the common practice \cite{han2021redet}, we use both the training set and validation set for training. We crop the original images into patches of size $\text{1024}\times \text{1024}$ with the stride of 824.

DOTA-v1.5 has the same images as DOTA-v1.0 but with different labels and more small targets (e.g., less than 10 pixels), and an additional category container crane. It has a total of 403,318 instances. \texttt{Container Crane} (CC)

For the multi-scale task, the images were rescaled with factors of \{0.5, 1.0, 1.5\} and cropped into patches of size $\text{1024}\times \text{1024}$ with a stride of 524. 

HRSC2016\cite{liu2017high} is an aerial image dataset for detecting ships labeled with oriented bounding boxes. It has 1,061 images with sizes ranging from 300×300 to 1500×900, divided into 436, 181, and 444 images for the training set, validation set, and testing set. We use both training set and validation set for training, and testing set for testing. We resized the images to $\text{800}\times\text{800}$.

\subsubsection{Implementation Details}
We implemented our method based on the widely-adopted MMRotate\cite{zhou2022mmrotate} library. In the pre-classification stage, we set $\sigma$ as $\frac{\sqrt{2}}{2}$. In the loss function, we set $\gamma$ as 0.5. Different baseline methods used different optimizers, which are detailed in the supplementary material. For each plug-and-play experiment, we trained with the settings reported in the original paper of the base model. We refer the readers to the \textit{Supplementary Materials} for further details.   


\input{tables/dota15_all}
\input{tables/hrsc_all}

\subsection{Improvements to Existing Methods}

We plugged \model into ReDet~\cite{han2021redet} to achieve SOTA on DOTA-v1.5 and HRSC2016. We show in \cref{DOTAv15_all} that our method achieves 77.59 mAP with multi-scale training and testing on the DOTA-v1.5 dataset, surpassing all previous methods by a clear margin. We also observe significant improvement when plugging \model into other methods.

We show in~\cref{HRSC2016_all} that our method achieves 90.70 mAP under VOC2007 metric and 98.62 mAP under VOC2012 metric on the HRSC2016 dataset, surpassing all other existing methods.

To further validate that \model is generally applicable, we add \model on top of different two-stage methods and report the detection results of experiments on DOTA-v1.0 in \cref{dotav10_plug} and on HRSC2016 in \cref{HRSC2016_plug}. We find that our method consistently improves the performances of various two-stage methods. Interestingly, we observe consistently larger improvements over average precisions (APs) at higher IoU thresholds. Since our method does not refine the regression head, this biased improvement suggests that our method is making better classifications on the proposals that were previously accurately located but misclassified. 

\input{tables/plugins}

\subsection{Experimental result on MS COCO}
\specialtext{To show that our method is generally applicable beyond the aerial image domain, we also conduct experiments on MS COCO. Using Faster R-CNN and Mask R-CNN as baselines, we find that \model provides consistent improvement without any further parameter tuning, as shown in \cref{tab:coco_result}. However, since MS COCO includes images of notably better quality and clarity compared to aerial images, the improvements afforded by \model are less significant than those on aerial object detection datasets like DOTA.}

\specialtext{We further validate the efficacy of \model by purposely blurring the images using a mean kernel. In this case, \model yields a more substantial improvement, which aligns with our motivation that the feedback information is helpful in response to image quality.}

\input{tables/coco}

\subsection{Recurrent Feedback using cascade \model}

\input{tables/cascade_flex}

\specialtext{The feedback mechanism in biological neural systems is believed to function in an iterative manner. Analogously, since \model is a plug-and-play module, we further explored plugging \model into itself to form a cascaded feedback structure. 
Specifically, we utilized the classification result output from the previous refinement layer as classification feedback for the subsequent refinement layer. The image feedback, like the original \model method, was obtained through the shared feature pyramid. Additionally, we observed that adding one layer only increases training time by approximately 5\%.}

\specialtext{As shown in \cref{Cascade}, additional \model layers can further improve performance by smaller margins, and the improvement wears out at 3 layers on DOTA-v1.5.}

\specialtext{The performance first increases and then decreases as the number of layers increases. This may be attributed to the classification feedback requiring high classification accuracy from the preceding layer to function optimally. When the number of layers increases, the variation in classification output from the initial layers during the early training epochs can hinder the training of subsequent layers. Only when the early layers stabilize can the later layers learn more meaningful information. Thus training cascade \model may require more than 12 epochs.}

\subsection{Ablation Studies}
To analyze and understand each component of our method, we use ReDet\cite{han2021redet} as the base method to conduct ablation studies on DOTA v1.5.

We first examine the effectiveness of each component in the feedback refine stage in \cref{feedback_type}. The column "multi-level" refers to fusing features with weights produced by a fixed Gaussian kernel as in the pre-classification stage. We observe that each component improves the final mAP, while the multi-level fusion contributes the most. 

We further examine the form of classification feedback in \cref{table:weight_parameterization}. The baseline method refers to the original method where features are extracted from a single feature level according to \ac{roi} area. Direct refers to generating $W^\text{cls}$ directly from \ac{roi} features. Gaussian is similar to Interpolation, our full model, except the generated values are a Gaussian kernel's mean and standard deviation. 

Both ablation studies show that fusing features from multiple levels bring the largest improvement and that incorporating feedback from image and classification further improves the performance. Furthermore, \cref{table:weight_parameterization} shows the importance of the linear interpolation we used to compute the fusion weights. 

\input{tables/ablation}

\subsection{Further Analyses}
\subsubsection{Parameterization of the Classification Feedback}
The most straightforward way to obtain the classification feedback $W^\text{cls}$ is to directly compute the weights for the N levels with a fully connected network. Although being the most flexible, this approach does not directly leverage the \ac{roi} area information. In contrast, a second option is to generate the parameters of a \textbf{Gaussian} kernel to compute the weights, similar to \cref{eq:gaussian_weight}. However, this approach may fall too restrictive as the optimal weight distribution may not be Gaussian. We argue that the \ac{roi} area is a valuable prior since the object size inherently implies its scale. Our proposed interpolation approach captures this information while allowing the weights to be more flexible. As shown in \cref{table:weight_parameterization}, \textbf{Interpolation} performs best. Comparing \textbf{Interpolation} and \textbf{Direct}, we find that incorporating \ac{roi} area can improve mAP by a clear margin, which is in line with our intuition. Moreover, we empirically observe that the learned classification feedback weights $W_k^{\text{cls}}$ roughly have a bell shape, which implies the Gaussian kernel is also a good prior.

\subsubsection{Image Feedback on Blurry Images}

\input{tables/image_quality}

The intuition that motivated this work is that the human brain adapts to blurry images, and so should neural networks. To verify if our method does adapt to images of low quality, we intentionally blurred clean images and observed how our method reacted to them. 

Specifically, we applied mean kernels with various sizes to images from DOTA-v1.5 and examined how the image feedback $\Phi^\text{img}$ responded to the manually imposed blurs in \cref{iq_mean}. We observe that the first feature level, which contains more image details, receives higher weights for cleaner images; and that the last level, which contains higher-level information, is more useful when dealing with blurrier images. This observation matches our intuition. 

We then selected the 10\% images with the largest $\Phi^\text{img}_1$ values from DOTA-v1.5 and applied mean kernels of different sizes. We show the observed changes in $\Phi^\text{img}$ in \cref{top10}. We find that these images show more significant changes in $\Phi^\text{img}_1$ and $\Phi^\text{img}_N$ than in \cref{iq_mean}. This also matches our expectation because, in \cref{iq_mean}, some images are already of low quality when all images are considered. Adding blur to low-quality images is less significant than applying the same blur to clean images. 

\begin{figure}
    \centering
    \includegraphics[width=\linewidth]{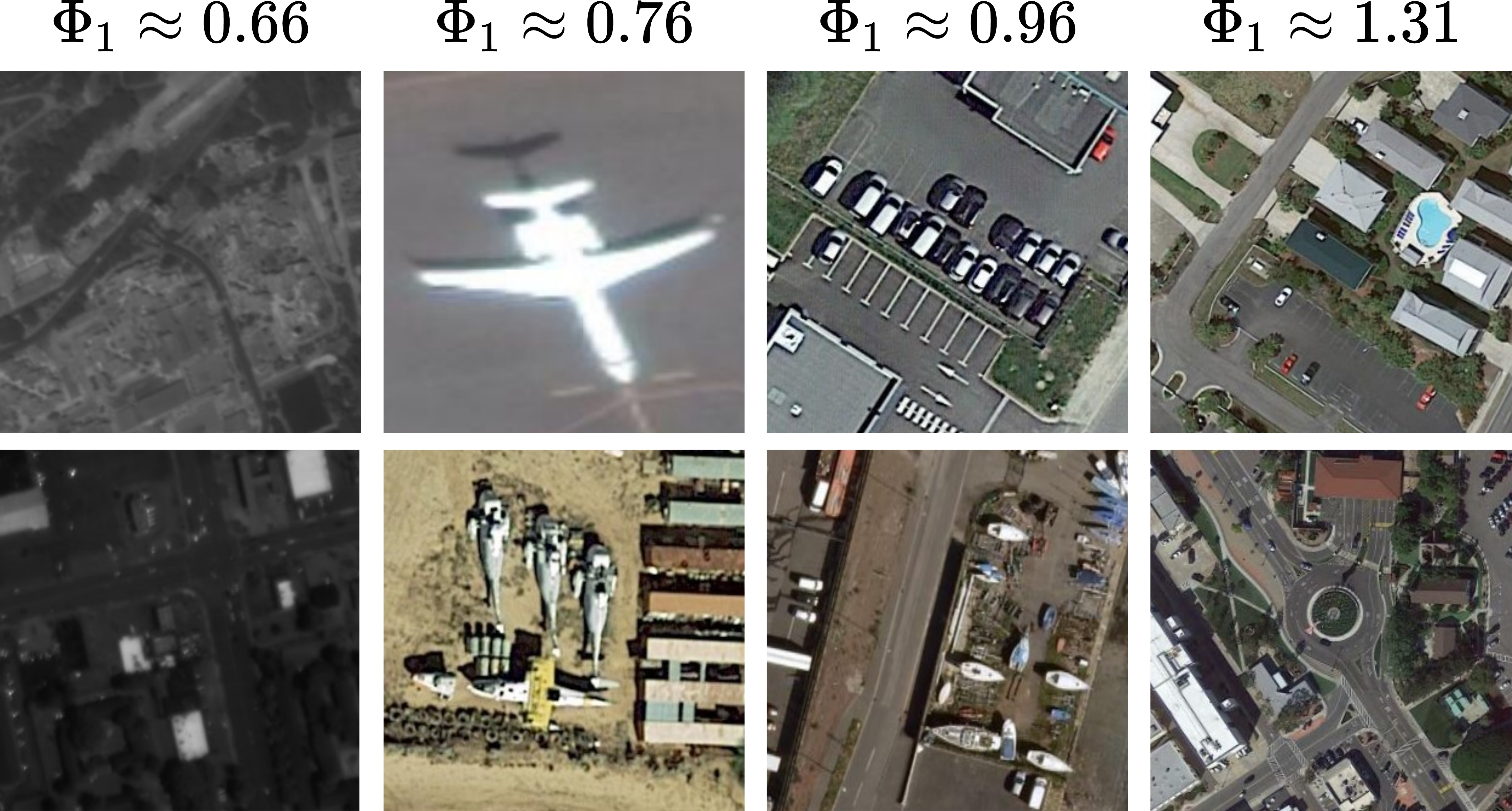}
    \caption{The correlation between image quality and the feature weight of the last level. $\Phi_1^\text{img}$ is generally large for images with better quality, e.g., better lighting, higher contrast, and clearer boundaries.}
    \label{fig:w1}
\end{figure}

\begin{figure}
    \centering
    \includegraphics[width=\linewidth]{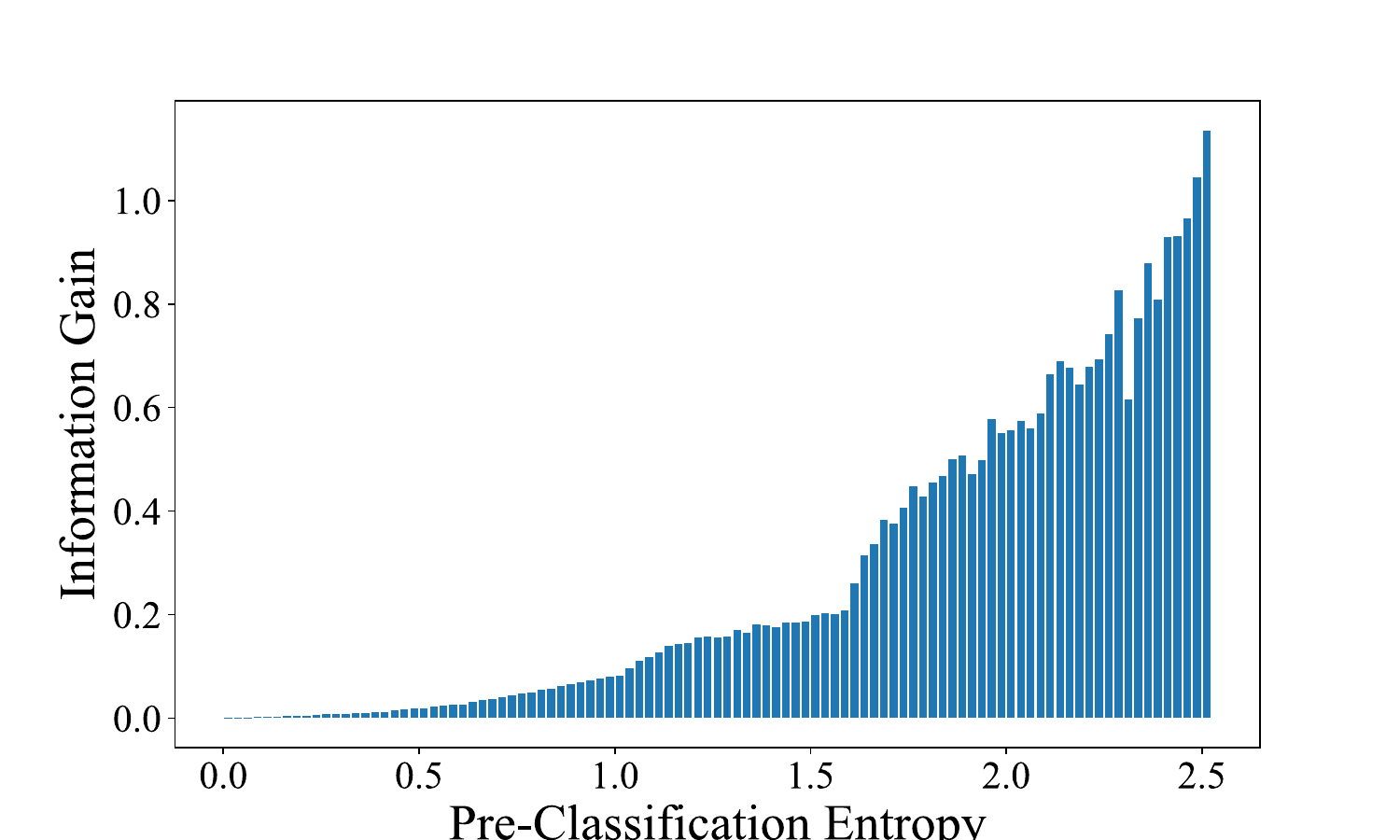}
    \caption{Information gain attained through the refinement stage. The $x$-axis is the entropy of the pre-classification result in bits. Higher entropy indicates uncertain class prediction, which is expected to be resolved with the feedback refinement process.}
    \label{fig:information_gain}
\end{figure}

Both analyses suggest that the image feedback learned to adapt to the blurs in the images to improve detection in low-quality images, which confirms our intuition. 

We further show image patches with different values of $\Phi_1^\text{img}$ in \cref{fig:w1}. We can observe a clear tendency that the image quality improves as values of $\Phi_1^\text{img}$ increase.

\begin{figure*}[htbp]
    \centering
    \includegraphics[width=0.90\textwidth]{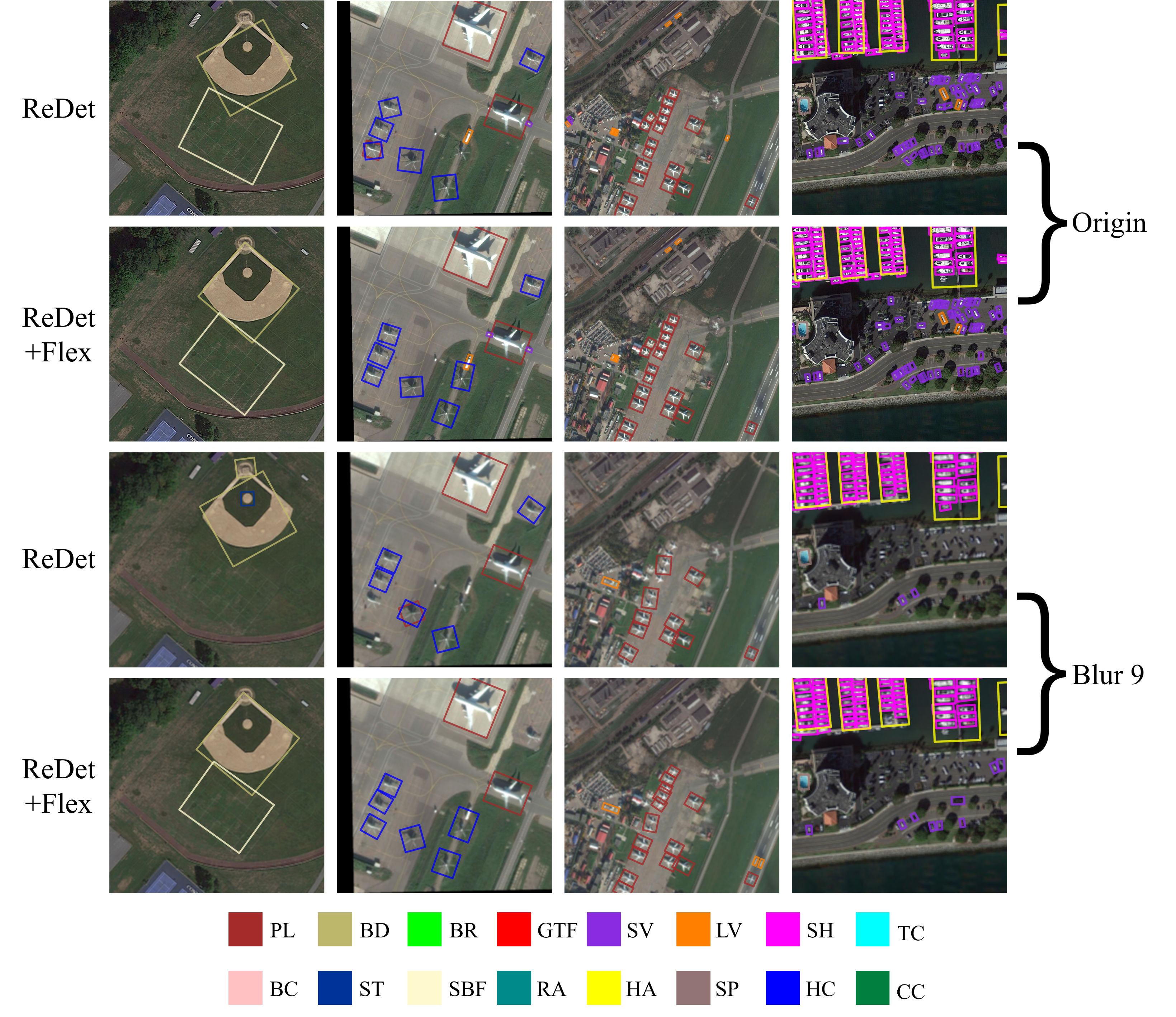}
    \caption{Examples of detection results of ReDet and ReDet+\model. }
    \label{fig:qualitative}
\end{figure*}

\subsubsection{Information Gain on Classification Result Feedback}

To inspect how the feedback benefits detection more closely, we examine the information gain attained with the refined features by computing the entropy difference between the classifications from the pre-classification $Y_\text{pre}$ and from the feedback refinement stages $Y_\text{refine}$. Specifically, it is calculated as:

\begin{equation}
    IG = H(Y_\text{pre}) - H(Y_\text{refine})
\end{equation}
where $H(X)=-\sum P(x)\log P(x)$ is the information entropy.

The correlation between the information gain and the entropy in the pre-classification stage is depicted in \cref{fig:information_gain} (in bits). We observe that for pre-classification results with larger entropy, \textit{i.e.}, higher uncertainty, the information gain is more substantial. More interestingly, we observe two sudden jumps at around $H(Y_\text{pre})=1.0$ and $H(Y_\text{pre})=1.6$. These two values coincide with $\log_2{2}$ and $\log_2{3}$, which account for the case where a classifier cannot identify an \ac{roi} between 2 and 3 similar classes, respectively. However, when the pre-classification entropy is low, the information gain is less significant, indicating that the refinement stage has less impact on more distinct classes that can already be predicted fairly well without feedback information.

\subsection{Limitations}
Although our model has shown significant improvement in average precision at higher IoUs, we did not refine the bounding box regression head for performance issues. We suspect the reason is that compared to the classification head that is sensitive to image qualities, the regression head performs fairly well in low-quality images. Therefore, refining the regression head does not bring much improvement. Another limitation of our submission is that we only experimented with refining only once, while it is entirely possible to refine for multiple iterations and attain larger improvements. 

%% file: tables/dota15_all.tex
\begin{table*}[] 
\centering
\caption{Results of each object class on the DOTA-v1.5 dataset. 
MS means multi-scale training and testing, and RR means random rotation augmentation. We highlight \model improvements in \highlight{blue}.}
\label{DOTAv15_all}
\begin{adjustbox}{max width=\textwidth}
\begin{tabular}{l|cc|cccccccccccccccc|c}
\toprule
Method                               & MS  & RR  & PL    & BD    & BR    & GTF   & SV    & LV    & SH    & TC    & BC    & ST    & SBF   & RA    & HA    & SP    & HC    & CC    & mAP   \\ \hline
RetinaNet-O\cite{lin2017focal}       &              &              & 71.43 & 77.64 & 42.12 & 64.65 & 44.53 & 56.79 & 73.31 & 90.84 & 76.02 & 59.96 & 46.95 & 69.24 & 59.65 & 64.52 & 48.06 & 0.83  & 59.16 \\
Mask R-CNN\cite{he2017mask}          &              &              & 76.84 & 73.51 & 49.90 & 57.80 & 51.31 & 71.34 & 79.75 & 90.46 & 74.21 & 66.07 & 46.21 & 70.61 & 63.07 & 64.46 & 57.81 & 9.42  & 62.67 \\
HTC \cite{chen2019hybrid}            &              &              & 77.80 & 73.67 & 51.40 & 63.99 & 51.54 & 73.31 & 80.31 & 90.48 & 75.21 & 67.34 & 48.51 & 70.63 & 64.84 & 64.48 & 55.87 & 5.15  & 63.40 \\
OWSR \cite{li2019learning}           & $\checkmark$ & $\checkmark$ & -     & -     & -     & -     & -     & -     & -     & -     & -     & -     & -     & -     & -     & -     & -     & -     & 74.90 \\ \hline
RF. R-CNN \cite{ren2015faster}        &              &              & 72.20 & 76.43 & 47.58 & 69.91 & 51.99 & 70.52 & 80.27 & 90.87  & 79.16 & 68.63 & 59.57 & 72.34 & 66.44 & 66.07 & 55.29 & 6.87 & 64.63 \\
RF. R-CNN + \model                   &              &              & 72.15 & \highlight{82.50} & \highlight{48.82} & \highlight{70.98} & 51.81 & \highlight{70.78} & 80.23 & 90.86 & 78.69 & 67.71 & 59.23 & \highlight{73.14} & 66.31 & \highlight{66.37} & \highlight{55.63} & \highlight{9.16} & \highlight{65.27}  \\ \hline
\ac{roi} Trans. \cite{ding2019learning}   &              &              & 72.27 & 81.95 & 54.47 & 70.02 & 52.49 & 76.31 & 81.03 & 90.90 & 84.19 & 69.12 & 62.85 & 72.73 & 68.67 & 65.89 & 57.09 & 7.12  & 66.69 \\
\ac{roi} Trans. + \model                    &              &              & \highlight{72.33} & \highlight{82.83} & \highlight{55.18} & \highlight{71.55} & 52.41 & \highlight{76.54} & \highlight{81.06} & 90.86 & 78.85 & 69.10 & \highlight{62.91} & 72.43 & \highlight{75.60} & \highlight{66.14} & \highlight{59.94} & \highlight{8.27}  & \highlight{67.25} \\ \hline
ReDet  \cite{han2021redet}           &              &              & 79.20 & 82.81 & 51.92 & 71.41 & 52.38 & 75.73 & 80.92 & 90.83 & 75.81 & 68.64 & 49.29 & 72.03 & 73.36 & 70.55 & 63.33 & 11.53 & 66.86 \\
ReDet + \model                         &              &              & \highlight{80.26} & 81.65 & \highlight{54.15} & \highlight{73.43} &\highlight{52.80} & \highlight{76.15} & \highlight{88.18} & \highlight{90.87} & \highlight{85.28} & \highlight{74.10} & \highlight{63.05} & \highlight{74.57} & \highlight{76.47} & \highlight{72.23} & \highlight{64.48} & \highlight{16.88} & \highlight{70.28} \\
ReDet + \model                         &              & $\checkmark$ & \highlight{80.62} & \highlight{84.78} & \highlight{55.08} & \highlight{72.49} & \highlight{57.95} & \highlight{80.86} & \highlight{88.56} & \highlight{90.88} & \highlight{85.73} & \highlight{73.80} & \highlight{63.73} & \highlight{73.55} & \highlight{76.56} & \highlight{71.91} & \highlight{68.78} & \highlight{20.59} & \highlight{71.62} \\ \hline
ReDet \cite{han2021redet}            & $\checkmark$ & $\checkmark$ & 88.51 & 86.45 & 61.23 & 81.20 & 67.60 & 83.65 & 90.00 & 90.86 & 84.30 & 75.33 & 71.49 & 72.06 & 78.32 & 74.73 & 76.10 & 46.98 & 76.80 \\
ReDet + \model                         & $\checkmark$ & $\checkmark$ & 86.52 & 85.03 & \highlight{61.27} & \highlight{81.73} & \highlight{67.96} & \highlight{83.70} & 89.81 & \highlight{90.88} & \highlight{86.88} & \highlight{84.10} & 71.43 & \highlight{76.90} & \highlight{78.44} & 74.18 & 73.36 & \highlight{49.20} & \highlight{77.59} \\ \bottomrule
\end{tabular}
\end{adjustbox}
\end{table*}

%% file: tables/hrsc_all.tex
\begin{table}[]
\caption{Results on the HRSC2016 dataset. We highlight the best result in \highlight{blue}.}
\label{HRSC2016_all}
\centering
\begin{tabular}{l|cc}
\toprule
Method          & mAP\textsubscript{07}  & mAP\textsubscript{12} \\ \hline
R2CNN           & 73.07      & 79.73      \\
Rotated RPN\cite{zhang2018toward}     & 79.08      & 85.64      \\
DRN\cite{pan2020dynamic} & -          & 92.70      \\
\ac{roi} Trans.      & 86.20      & -          \\
Gliding Vertex\cite{xu2020gliding}  & 88.20     & -          \\
R3Det\cite{yang2021r3det}   & 89.26      & 96.01      \\
S2A-Net\cite{han2021align}  & 90.17      & 95.01      \\
ReDet           & 90.46      & 97.63      \\
Oriented RCNN\cite{xie2021oriented}   & 90.50      & 97.60      \\
ReDet + \model    & \highlight{90.70}      & \highlight{98.62}      \\ \bottomrule
\end{tabular}
\end{table}

%% file: tables/plugins.tex
\begin{table}[]
\small
\caption{Results in COCO style on the DOTA-v1.0 dataset. * means multi-scale training and testing with random rotation augmentation. For ReDet*, it didn't report AP\textsubscript{75} and mAP in paper. We highlight \model improvements in \highlight{blue}.}
\label{dotav10_plug}
\centering
\begin{tabular}{l|lll}
\toprule
Method                      & AP\textsubscript{50} & AP\textsubscript{75} & mAP          \\ \hline
RF. R-CNN        & 73.40                & 39.61                & 40.76        \\
with \model & \highlight{73.95(+0.55})         & \highlight{40.77(+1.16)}         &\highlight{41.66(+0.90)} \\ \hline
ReDet                       & 76.25                & 50.86                & 47.11        \\
with \model               & \highlight{77.09(+0.84) }        & \highlight{51.90(+1.04) }        & \highlight{48.70(+1.59)} \\ \hline
ReDet*                      & 80.10                & -                    & -            \\
with \model               &\highlight{80.61(+0.51) }        & \highlight{58.87}                & \highlight{53.85}        \\ \bottomrule
\end{tabular}
\end{table}

\begin{table}[] 
\small
\centering
\caption{Results in COCO style on the HRSC2016 dataset. \ddag  means using random rotation augmentation. For Gliding Vertex and \ac{roi} Transformer, they didn't report AP\textsubscript{75} and mAP in paper. We highlight \model improvements in \highlight{blue}.}
\label{HRSC2016_plug}
\begin{tabular}{l|lll}
\toprule
Method                      & AP\textsubscript{50}         & AP\textsubscript{75}          & mAP          \\ \hline
RF. R-CNN        & 84.97        & 35.89         & 41.80        \\
with \model & \highlight{86.85(+1.88)} & \highlight{47.75(+11.86)} & \highlight{46.16(+4.36)} \\ \hline
Gliding V.              & 88.20        & -             & -            \\
with \model       & \highlight{89.19(+0.99)} & \highlight{67.85}         & \highlight{58.11}        \\ \hline
\ac{roi} Trans.             & 86.20        & -             & -            \\
with \model     & \highlight{90.04(+3.84)} & \highlight{78.96}         & \highlight{64.50}        \\ \hline
ReDet                       & 90.46        & 89.46         & 70.41        \\
with \model                & \highlight{90.58(+0.12)} & \highlight{89.99(+0.46)}  & \highlight{74.23(+3.82)} \\
with \model\ddag               & \highlight{90.70(+0.24)} & \highlight{90.03(+0.57)}  & \highlight{76.43(+6.02)} \\ \bottomrule
\end{tabular}
\end{table}

%% file: tables/coco.tex
\begin{table}[]
\centering
\caption{Results on MS COCO dataset. FR. means Faster R-CNN\cite{ren2015faster}, and MR. means Mask R-CNN\cite{he2017mask}. "Blur X" means a $\text{X}\times \text{X}$ kernel is used to blur the images. We highlight \model improvements in \highlight{blue}.}
\label{tab:coco_result}
\begin{tabular}{l|lll}
\toprule
\multicolumn{1}{l|}{\multirow{2}{*}{Method}} & \multicolumn{3}{c}{mAP}                                                               \\ \cline{2-4} 
\multicolumn{1}{c|}{}                        & \multicolumn{1}{c}{Origin} & \multicolumn{1}{c}{Blur 9} & \multicolumn{1}{c}{Blur 15} \\ \hline
FR. \cite{ren2015faster}                                           & 37.7                       & 29.8                       & 24.6                        \\
with \model                                      & \highlight{38.0 (+0.3)}                & \highlight{30.7 (+0.9)}                & \highlight{25.7 (+1.1)}                 \\ \hline
MR. \cite{he2017mask}                                           & 38.5                       & 30.6                       & 25.5                        \\
with \model                                      & \highlight{38.8 (+0.3)}                & \highlight{31.6 (+1.0)}                & \highlight{26.5 (+1.0)}                 \\ \bottomrule
\end{tabular}
\end{table}

%% file: tables/cascade_flex.tex
\begin{table*}[t]
\centering
\caption{Results on DOTA-v1.5 of Cascading \model using ReDet as baseline.}
\label{Cascade}
\vspace{6pt}
\begin{tabular}{c|cccccc}
\hline
\# Refine Layers & 0 & 1 & 2 & \highlight{3} & 4 & 5   \\ \hline
mAP & 66.86 & 70.28 & 70.59 & \highlight{70.78} & 70.34 & 70.13 \\ \hline

\end{tabular}
\end{table*}

%% file: tables/ablation.tex
\begin{table}[]
\centering
\begin{minipage}{0.45\textwidth}
\centering
\caption{Ablation study of each component in the feedback refine stage.}
\label{feedback_type}
\resizebox{\columnwidth}{!}{
\begin{tabular}{ccc|c}
\toprule
Multi-level  & Class feedback & Image feedback  & mAP   \\ \hline
             &              &              & 66.86 \\
$\checkmark$ &              &              & 69.13 \\
$\checkmark$ & $\checkmark$ &              & 69.88 \\
$\checkmark$ & $\checkmark$ & $\checkmark$ & 70.28 \\ \bottomrule
\end{tabular}
}
\end{minipage}\hfill
\vspace{6pt}
\begin{minipage}{0.45\textwidth}
\centering
\caption{Ablation study of different learning methods of pre-classification feedback. For the \textbf{Input} column, it regards what we use for computing the weights, and \textbf{A} means we use the RoI area, and \textbf{C} means we use the classification feedback.}
\label{table:weight_parameterization}
\resizebox{\columnwidth}{!}{
\begin{tabular}{c|c|c|c}
\toprule
Learning Method   & Input & Parameterization & mAP   \\ \hline
Baseline      & A     & Fixed            & 66.86 \\
Direct        & C     & Learned          & 69.78 \\
Gaussian      & C+A   & Gaussian         & 69.60 \\
Interpolation & C+A   & Learned          & 70.28 \\ \bottomrule
\end{tabular}
\textbf{}}
\end{minipage}
\end{table}

%% file: tables/image_quality.tex

\begin{table}[]
\centering
\caption{Average image quality feedback on clean and blurred images from DOTA-v1.5. Blur X means applying $\text{X}\times \text{X}$ mean kernel.}
\label{iq_mean}
\begin{tabular}{l|cc}
\toprule
Input  & First layer & Last layer \\ \hline
Origin & 0.8289      & 1.1806     \\
Blur 5  & 0.8175      & 1.1855     \\
Blur 9  & 0.7970      & 1.2050     \\
Blur 21 & 0.7768      & 1.2223     \\ \bottomrule
\end{tabular}
\end{table}
\begin{table}[]
\centering
\caption{Average image quality feedback on the top 10\% clean images and their blurred versions from DOTA-v1.5. Blur X means applying $\text{X}\times \text{X}$ mean kernel. The last column shows the average of $W_N$ over the dataset. }
\label{top10}
\begin{tabular}{l|ccc}
\hline
Input   & First layer & Last layer & Last layer avg.\\ \hline
Origin  & 1.1477      & 0.9897     & 1.1806\\
Blur 5  & 1.0955      & 1.0166     & 1.1855\\
Blur 9  & 0.9910      & 1.0768     & 1.2050\\
Blur 21 & 0.8490      & 1.1689     & 1.2223\\ \bottomrule
\end{tabular}
\end{table}

%% file: sec/5_conclusion.tex
\section{Conclusion}
We proposed a feedback-based \ac{roi} feature extraction method that can be easily plugged into any two-stage detectors that extract features from multi-level feature maps. Specifically, this method divides the extraction of \ac{roi} into a pre-classification stage and a feedback refine stage. The refinement stage considers feedback from both the image feature and the \ac{roi} classification from the pre-classification stage and adaptively adjusts the weights of the features fused from different layers. Extensive experiments on DOTA and HRSC2016 show that our method achieves consistent improvements over various baseline methods, including the SOTA, and further analyses show that the improvements are closely related to image quality, which aligns with our motivation. 